\documentclass[pmlr]{jmlr}

\jmlrproceedings{}{}
\jmlrvolume{}
\jmlryear{}
\jmlrpages{}

\RequirePackage{graphicx}
\usepackage{siunitx}
\usepackage{array}
\usepackage{booktabs}
\usepackage{longtable}
\usepackage{multirow}
\makeatletter
\def\set@curr@file#1{\def\@curr@file{#1}} 
\makeatother
\usepackage[load-configurations=version-1]{siunitx} 
\theorembodyfont{\upshape}
\theoremheaderfont{\scshape}
\theorempostheader{:}
\theoremsep{\newline}

\title[]{Fine-Tune, Don't Prompt, Your Language Model to Identify Biased Language in Clinical Notes}

\author{\begin{center}
\normalfont Isotta Landi\textsuperscript{1$\ast$} \quad Eugenia Alleva\textsuperscript{1$\ast$}\\
Nicole Bussola\textsuperscript{1} \quad Rebecca M. Cohen\textsuperscript{2} \quad
Sarah Nowlin\textsuperscript{3}\\ Leslee J. Shaw\textsuperscript{4}\quad Alexander W. Charney\textsuperscript{1}\quad Kimberly B. Glazer\textsuperscript{5}
\end{center}}

\makeatletter
\renewcommand{\@titlefoot}{}
\makeatother

\begin{document}
\maketitle

\bgroup
\let\oldthefootnote\thefootnote
\renewcommand{\thefootnote}{}
\let\oldfootnoteseptext\footnoteseptext
\renewcommand{\footnoteseptext}{}
\footnotetext{%
$^\ast$Equal contribution. 
\textsuperscript{1}Department of AI and Human Health, Icahn School of Medicine at Mount Sinai, NY. 
\textsuperscript{2}Icahn School of Medicine at Mount Sinai, NY. 
\textsuperscript{3}Population Health Science and Policy, Icahn School of Medicine at Mount Sinai, NY.
\textsuperscript{4}Obstetrics, Gynecology and Reproductive Science, Blavatnik Family Women’s Health Research Institute, Icahn School of Medicine at Mount Sinai, NY.
\textsuperscript{5}Obstetrics and Gynecology, University of Pennsylvania, PA.
Correspondence to: Isotta Landi \textless landi.isotta@gmail.com\textgreater, Eugenia Alleva \textless eugeniaalessandrae.allevabonomi@mssm.edu\textgreater.
}
\egroup

\setcounter{footnote}{0}
\renewcommand{\thefootnote}{\arabic{footnote}}

\begin{abstract}
Clinical documentation can contain emotionally charged language with stigmatizing or privileging valences. We present a framework for detecting and classifying such language in clinical text as \emph{stigmatizing}, \emph{privileging}, or \emph{neutral}. First, we constructed a curated lexicon of biased terms, scoring each for emotional valence. We then used lexicon-based matching to extract text chunks from OB-GYN delivery notes (Mount Sinai Hospital, New York) and MIMIC-IV discharge summaries across multiple specialties. Three clinicians independently annotated all chunks, enabling characterization of valence patterns across specialties and healthcare systems.

We benchmarked multiple classification strategies, namely, zero-shot prompting, in-context learning, and supervised fine-tuning, across encoder-only models (GatorTron) and generative large language models (Llama). Fine-tuning with lexically primed inputs consistently outperformed prompting approaches. The clinical language model GatorTron achieved an $F_1$ score of $0.96$ on the OB-GYN test set, outperforming larger generative models while requiring minimal prompt engineering and fewer computational resources. External validation on MIMIC-IV revealed limited cross-domain generalizability ($F_1 < 0.70$, 44\% drop). Training on the broader MIMIC-IV dataset improved generalizability when testing on OB-GYN ($F_1 = 0.71$, 11\% drop), but at the cost of reduced precision. 

Our findings demonstrate that fine-tuning outperforms prompting for emotional valence classification and that models must be adapted to specific medical specialties and institutional contexts to achieve clinically appropriate performance. As such, the same terms can carry different emotional valences across specialties: words with clinical meaning in one context may be used in stigmatizing ways in another. For bias detection, where misclassification risks either undermine clinician trust or perpetuate patient harm, specialty-specific fine-tuning is essential to reliably capture these semantic shifts. 

\end{abstract}

\section{Introduction}
Improving the characterization and extraction of biased language in clinical notes is a critical priority for both machine learning and healthcare research. If we consider the clinical text as a narrative of the patient-clinician interaction, the investigation of the emotional valence of the language used can reveal signs of strained therapeutic relationships. By emotional valence, we refer to the positive, negative, or neutral emotional response elicited by a text, an aspect that has been linked to implicit biases held by clinicians toward minoritized and historically marginalized groups \citep{sun2022, martin2020}. Because these biases can adversely affect the quality of care delivered \citep{goddu2018}, their identification can inform public policy and support the design of targeted interventions that promote more equitable healthcare practices. Moreover, as large language models (LLMs) move closer to real-world deployment as clinical support tools \citep{hager2024}, it becomes essential to reliably extract biased language at scale from pretraining and fine-tuning datasets to prevent its amplification \citep{tate2024} and its negative downstream effects on the clinicians who read the model-generated documentation \citep{goddu2018}.

Addressing this challenge is complicated by the lack of consensus on how to define biased language and standardize annotation practices. Bias has been characterized in various ways, e.g., as a positive/negative emotional valence \citep{sun2022} or as linguistic microaggressions \citep{harrigian2023}. Furthermore, approaches to studying biased language vary widely: some rely directly on the manual annotation of real-world clinical texts \citep{scroggins2025, park2021}, others use controlled vignette-based questionnaires \citep{goddu2018}, and some begin by curating vocabularies of relevant descriptors before extracting them from clinical narratives \citep{apakama2024, bilotta2024, boley2024, weiner2023, harrigian2023, himmelstein2022, sun2022}.

Although natural language processing (NLP) techniques—particularly LLMs—hold significant promise for this task \citep{pierson2025, barcelona2024}, there remains a lack of comprehensive research on best practices for (1) reliable and scalable extraction of biased text; (2) minimal annotation effort; and (3) generalization across clinical specialties and healthcare systems. Existing studies employ a wide range of NLP methods to identify biased language in clinical text, from rule-based extraction and word count approaches \citep{weiner2023, himmelstein2022, sun2022} to zero-shot prompting of LLMs for sentence classification \citep{apakama2024}. While encoder-based models pretrained on clinical text, such as ClinicalBERT \citep{alsentzer2019}, have demonstrated superior performance over traditional machine learning methods in classifying stigmatizing versus privileging language \citep{scroggins2025}, it is still unclear whether domain-specific knowledge is necessary for optimal LLM performance, and whether encoder-based pretrained language models (PLMs) offer distinct advantages over generative LLMs \citep{harrigian2023}.

In this study, we propose a multi-step framework for the characterization and extraction of biased language in clinical text operationalized as emotional valence. Our goal is to investigate how such valence varies across medical specialties and hospital systems \citep{harrigian2023} and to evaluate which NLP-based information extraction methods leveraging language models are most effective in capturing it. We started with the quantification of the stigmatizing or privileging valence of a curated lexicon taken from the literature. The most polarized terms were then used to extract relevant text segments centered on those keywords from OB-GYN delivery notes from the Mount Sinai Health system (NY) and MIMIC-IV hospital and emergency department discharge notes from the Beth Israel Deaconess Medical Center in Boston, USA \citep{johnson2023}. 

These anchored chunks were subsequently annotated by clinicians as \emph{stigmatizing}, \emph{privileging}, or \emph{neutral}. This three-class configuration simplifies the annotation process by grounding it in clinicians' emotional interpretation of a text, making it more intuitive, while also being general enough to enable the analysis of how medical specialties influence labeling variability across healthcare systems. 

PLMs and LLMs performance in classifying the emotional valence attributed to the clinical text was then evaluated and compared, leveraging different classification strategies, i.e., supervised fine-tuning, zero-shot inference, and in-context learning (ICL). To assess the impact of distribution shifts across healthcare systems and clinical specialties, we externally validated the models on the MIMIC-IV annotated text.

To facilitate research on patient-clinician communication, we make available our tuned models, de-identified annotated datasets (OB-GYN and MIMIC-IV), and valence-scored lexicon upon request to EA and execution of a data sharing agreement. Code can be accessed at \url{https://github.com/landiisotta/medsplex}.


    


\section{Related Work}
Our work builds on prior research in machine learning for healthcare, contributing important advances to the study of biased language. \cite{martin2020} were the first to operationalize bias in psychiatric nursing notes through emotional tone scoring of clinical text. We extended this approach by introducing a two-step annotation process, where the scoring of stigmatizing or privileging valence only happens on the vocabulary extracted from the literature. The annotation of the clinical text, segmented into chunks anchored in the keywords, is then limited to three sentiment classes, i.e., \emph{stigmatizing}, \emph{privileging}, or \emph{neutral}. This design improves scalability and interpretability by reducing cognitive load on annotators, who are asked to assess their emotional response to targeted text chunks. By annotating two independent datasets of non-overlapping specialties, i.e., obstetrics and gynecology care vs hospital and emergency departments, we also enabled a data-driven analysis of how emotional valence varies in different clinical contexts.

Previous work has explored zero-shot prompting with generative LLMs \citep{apakama2024} and supervised fine-tuning with encoder-only PLMs \citep{harrigian2023} for classifying biased language in clinical notes. In the general domain, the advantages of fine-tuning ``smaller'' encoder-based models over zero-shot approaches with large generative models for sentence classification tasks have been reported \citep{bucher2024}. However, existing work lacks a direct and comprehensive comparison of language models and tuning strategies for bias detection in the clinical domain, particularly with respect to understanding why certain approaches outperform others. \cite{harrigian2023} conducted the most in-depth study to date, evaluating NLP methods for classifying microaggressions in clinical notes across clinical specialties and hospital systems. They showed that BERT \citep{devlin2019} and ClinicalBERT \citep{alsentzer2019} outperformed traditional NLP approaches across three classification tasks, i.e., credibility/obstinacy, compliance, and descriptors. However, the contribution of medical knowledge on model performance remains unclear. Additionally, although model performance declined when tested on data from a different hospital system, insights into specialty-specific variation in bias perception were limited due to overlapping domains in the datasets. Our work advances these efforts by systematically evaluating generative and non-generative tuning strategies, with models pretrained on either clinical or general-domain text. We also demonstrated the importance of lexical priming and domain-specific adaptation.

\section{Methods}
\label{sec:methods}
\begin{figure}[t]
   \centering 
   \includegraphics[width=6in]{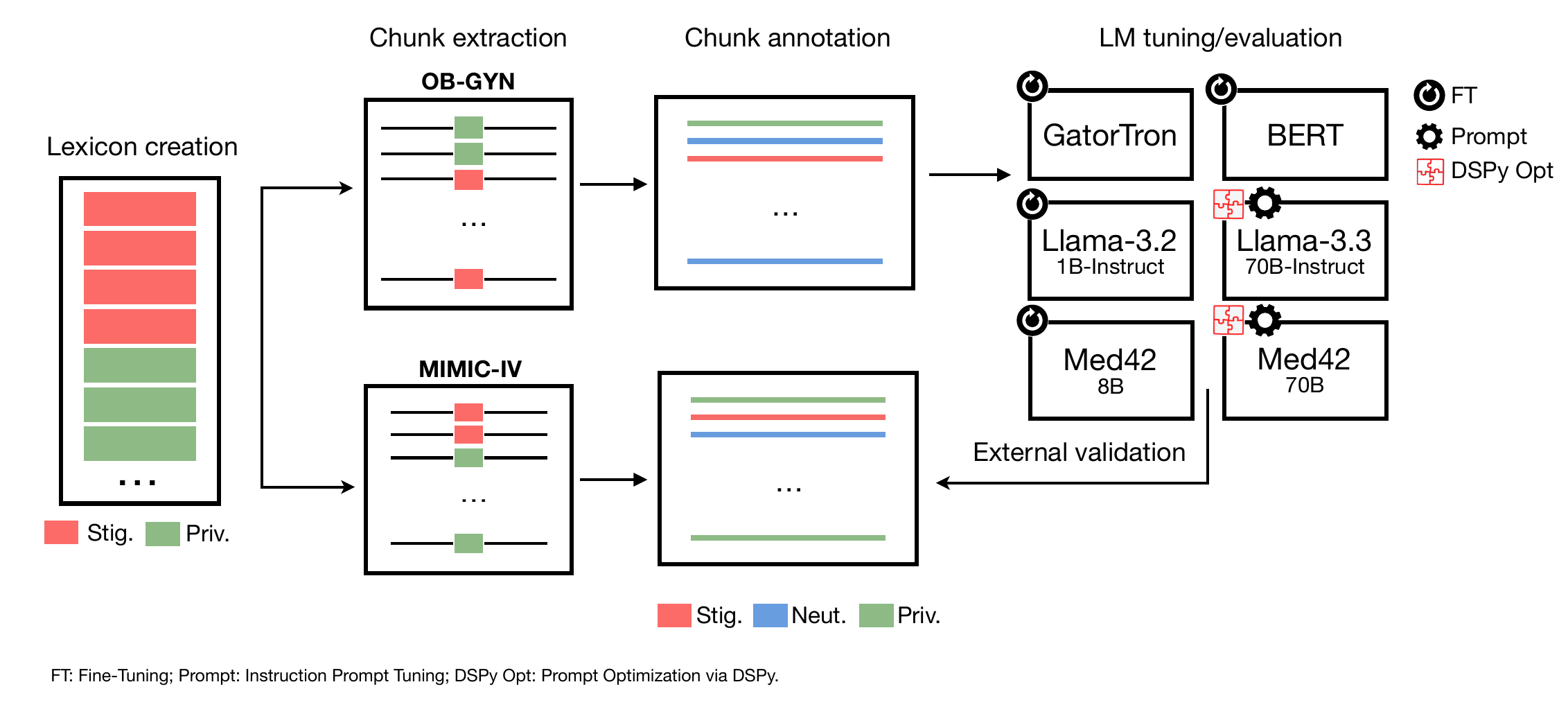} 
   \caption{Multi-stage methodology outline.}
   \label{fig:pipeline}
 \end{figure}

We designed the multi-stage methodology showcased in Figure~\ref{fig:pipeline} to include (1) a characterization of clinical text emotional valence in specific medical specialties; (2) the identification of the best NLP tools (either encoder-only PLMs or autoregressive LLMs) for the classification of biased language; (3) the assessment of whether emotional valence understanding and/or clinical knowledge are necessary for language models to perform the task; (4) the investigation of the generalization of such models to different clinical specialties and medical systems. We performed most computations on a single NVIDIA H100 GPU (80GB) using Python 3.10, PyTorch 2.5.1, and CUDA 12.4.

\subsection{Lexicon Creation}
We compiled a list of stigmatizing and/or privileging terms from the literature \citep{barcelona2023, sun2022, himmelstein2022, beach2021, fernandez2021, park2021}. We excluded terms that target specific groups of people based on sociodemographic characteristics, as those were not relevant to our analysis. This process resulted in an initial list of $62$ stigmatizing and/or privileging terms. Terms with both privileging and stigmatizing valences, e.g., \emph{compliant} and \emph{noncompliant}, were counted once.

We shared this list of terms, along with useful synthetic examples (see Table~\ref{tab:a1} in Appendix A), with four medical domain experts, including one maternal-fetal medicine fellow, one nurse practitioner, one nursing Ph.D. student, and one medical student. After being trained on the task, each expert rated the terms on a Likert scale from $1$ to $5$, where $1$ indicated ``I strongly agree this term is stigmatizing/privileging'' and $5$ indicated ``I strongly disagree this term is stigmatizing/privileging''. 

Each annotator rated all the terms twice, for privileging and stigmatizing valences, respectively, in a full-overlap setting. To compute inter-annotator agreement, we merged ratings of $1$ and $2$ into a single ``agree'' category, ratings of $4$ and $5$ into a ``disagree'' category, and ratings of $3$ into an ``undecided'' category. Percent agreement among annotators was lower for privileging ($47.79\%$) than for stigmatizing ($80.43\%$) valence. Gwet AC2 coefficient was $0.3970$ for privileging and $0.7736$ for stigmatizing valence, indicating moderate and substantial agreement, respectively. Terms with an average stigmatizing or privileging rating $\leq 2.5$ were included in the final list, which consisted of $55$ terms.

\subsection{Chunk extraction}
\paragraph{OB-GYN} We extracted $225,103$ clinical notes associated with $17,978$ newborn delivery episodes and authored by a physician, resident, or fellow at the Mount Sinai Healthcare system in New York. For each note, we extracted text chunks centered on the keywords from the lexicon with a $200$ character window. We performed keyword matching using rule-based regular expressions (e.g., \texttt{\textbackslash b(?:noncompliant|noncompliance|poor compliance|compliant)\textbackslash b} for the keyword \emph{compliance}). 

Out of the $55$ keywords, $42$ matched across all delivery notes. We randomly sampled a total of $1,015$ chunks from $889$ clinical notes. We observed a median of $1$ chunk per note (min/max: $1/7$) and a median of $5.5$ keyword matches across the dataset (min/max: $1/127$).

\paragraph{MIMIC-IV} We applied the same approach to create an external dataset using MIMIC-IV discharge summaries from inpatient and emergency department encounters at Beth Israel Deaconess Medical Center in Boston. All $55$ keywords successfully matched, and $5$ chunks per keyword were randomly selected, yielding a total of $275$ chunks from $274$ clinical notes.

In both cases, we did not apply any exclusion criteria to the keywords or the extracted chunks during text selection. With this inclusive approach, we aimed to capture the full range of contexts in which the keywords appear, including administrative text or templates, in order to investigate the emotional valence of real-world clinical text as comprehensively as possible.

\subsection{Chunk Annotation}
First, we manually removed protected health information (PHI) from OB-GYN clinical text chunks, following the Health Insurance Portability and Accountability Act Privacy Rule\footnote{\url{https://www.hhs.gov/hipaa/for-professionals/special-topics/de-identification}}. Two annotators with clinical expertise independently de-identified the full dataset (i.e., with complete overlap). The annotation efforts returned a negligible percentage difference of $0.01$ in annotated content. The dataset with the highest proportion of PHI, representing $2.46\%$ of the total character count, was selected, and all identified PHI was replaced with three underscores (“$\_\_\_$”), consistent with the de-identification approach used in MIMIC-IV.

Three annotators with clinical expertise then proceeded to annotate the full set of text chunks, from both the OB-GYN and MIMIC-IV datasets, with an overlap of $100$ chunks randomly extracted from the OB-GYN dataset, stratifying on anchor terms. The annotation agreement among annotators was high, Gwet AC2 agreement coefficient $0.8589$. 

The annotators were asked to label each chunk as \emph{stigmatizing}, \emph{privileging}, or \emph{neutral} and were instructed as follows: ``You are asked to label clinical note chunks, containing a highlighted term from a curated lexicon, as one of the following: (1) \emph{Stigmatizing}, if the term is used in a way that reflects negative judgment, bias, or devaluation of the patient; (2) \emph{Privileging}, if the term is used in a positive way, regardless of whether it might or might not convey positive bias, preferential treatment, or unjustified praise of the patient; (3) \emph{Neutral}, if the term refers to someone other than the patient (e.g., family member, clinician), is not referring to any person, but to a concept, action, or general advice, or appears as part of template or standard text''. See Table~\ref{tab:a2} in Appendix A for examples of annotated chunks.

The OB-GYN dataset was subsequently split into training ($60\%$), development ($20\%$), and test ($20\%$) sets. The annotated MIMIC-IV dataset was held out as an external validation set. Both the de-identification and annotation processes were performed using the Prodigy Annotation Tool\footnote{\url{https://prodi.gy/}}.

\subsection{Language Models Tuning Strategies}
From the Hugging Face platform, we considered BERT-large ($340M$ parameters) and GatorTron-base ($345M$ parameters), as encoder-only language models pretrained on general-domain and clinical text, respectively \citep{devlin2019, yang2022}. As autoregressive, instruction-tuned LLMs, we included Llama 3.3 ($70B$ parameters) and its smaller variant, Llama 3.2 ($1B$ parameters). We also considered two versions of Llama-2 models pre-trained on clinical text, i.e., Med42-8B and Med42-70B \citep{christophe2024}.

We assessed the generative capabilities of the Llama models in zero-shot, prompting, and fine-tuning settings. Generated outputs were post-processed and mapped to the three target labels: \emph{stigmatizing}, \emph{privileging}, and \emph{neutral}. BERT and GatorTron were fine-tuned both in a traditional setting, where the \texttt{[CLS]} token was used for classification with three classes, or a prompt-based fine-tuning approach with a cloze-style prompt. 

To investigate how models adapt to word usage and tone, input to the tasks were either the chunks anchored in the keywords or lexically primed inputs, where each corresponding keyword was added in front or at the end of the anchored text, e.g., \texttt{"Word: \{word\} [$\dots$] \{anchored chunk\} [$\dots$]"}.

\paragraph{Encoder-based Models}
We evaluated BERT and GatorTron models across three experimental settings for the classification of the emotional valence of clinical text snippets. First, we conducted standard fine-tuning by appending a classification head to the final \texttt{[CLS]} token representation and training the model directly on the labeled data.

Second, we implemented a prompt-based fine-tuning approach using cloze-style prompting. In this case, each input sentence was modified into: \texttt{"\{anchored text\} This sentence is: [MASK]"}. We employed manual verbalizers to map the output logits of the masked token to the three classes. Two verbalizer configurations were tested: (1) simple verbalizer using single words and (2) multi-word verbalizer using multiple representative tokens per class (see Table~\ref{tab:b1} in Appendix B). For each example, the model’s logits corresponding to the verbalizer tokens associated with each class were averaged and normalized. We then applied a softmax function to produce the final probability distribution over the three classes.

Third, to adapt the models to word usage in the OB-GYN clinical specialty, we leveraged the keywords from the lexicon to provide additional semantic cues to guide classification. We extended the prompting strategy with lexical priming by introducing keywords into the prompts. Sentences were reformulated as: \texttt{"\{anchored chunk\} Keyword is: \{word\}. This sentence is: [MASK]"}.

We performed hyperparameter optimization on the held-out development set varying the learning rate, verbalizer type, and batch size (see Appendix B for details). We used the macro-averaged $F_1$ score as the primary evaluation metric. We based early stopping and model selection on the $F_1$ score obtained on the development set. Final evaluation was performed on the held-out test set and the external MIMIC-IV dataset via bootstrapping, repeated 1,000 times.

\paragraph{Generative Models}
We first instructed Llama 3.3, a 70-billion-parameter language model fine-tuned on instructions, and Med42-70B, in a zero-shot setting to output the emotional valence label of the clinical text chunks. The task structure was defined using the DSPy framework \citep{khattab2023}, which enables replacing prompt engineering with structured and declarative natural-language modules. We started by specifying the input/output behavior of our model as a signature and selecting a prediction module for invoking the model. Two input configurations were tested to establish baseline performance: (1) lexically primed inputs to assess the impact of keyword priming on model performance (i.e., \texttt{"\{Instruction\}. Word: \{keyword\} [$\dots$] \{anchored chunk\} [$\dots$]}") and (2) standard instructions without lexical priming (see Table~\ref{tab:c1} in Appendix C).

To enhance the model's performance without fine-tuning it, we employed DSPy's MIPROv2-Multiprompt Instruction PRoposal Optimizer Version 2 \citep{opsahl2024}. MIPROv2 jointly optimizes task instructions and few-shot examples by (1) bootstrapping few-shot example candidates from the training set; (2) generating candidate instructions invoking the LLM itself; (3) identifying the optimal combination of instructions and examples via Bayesian optimization. The combinations were iteratively evaluated  on mini-batches from the development set using exact matching. Details on the MIPROv2 configurations explored, along with the best hyperparameters selected are provided in Appendix C.

Following prompt optimization, we conducted instruction fine-tuning experiments using Llama 3.2, a 1-billion-parameter instruction-tuned model and the Llama-2 based clinical version Med42-8B. The fine-tuning process involved: (1) data contextualization, where the model was grounded in OB-GYN clinical context to enhance domain-specific understanding using the instruction and examples from the MIPROv2 optimized prompt and (2) general instruction fine-tuning, where only the optimized instruction, without few-shot examples, was used to guide the fine-tuning process. See details on the hyperparameters explored in Appendix C.

We evaluated the models in a zero-shot setting directly on the development and test sets considered together. We performed optimization and instruction fine-tuning on the OB-GYN training set and evaluation on the development set from the same clinical setting for hyperparameter selection. The best models were tested on the OB-GYN test set and externally validated on the MIMIC-IV dataset. All evaluations were done using the macro-averaged $F_1$ score. To estimate performance distributions, we applied bootstrapping with 1,000 repetitions on the test and external validation sets.

\section{Results}
In the following, we report the results of our multi-stage methodology, addressing the research questions outlined at the beginning of Section~\ref{sec:methods}.

\subsection{Emotional Valence in Clinical Text Varies by Medical Specialty}
\label{sec:results1}
\begin{figure}[t]
   \centering 
   \includegraphics[width=6in]{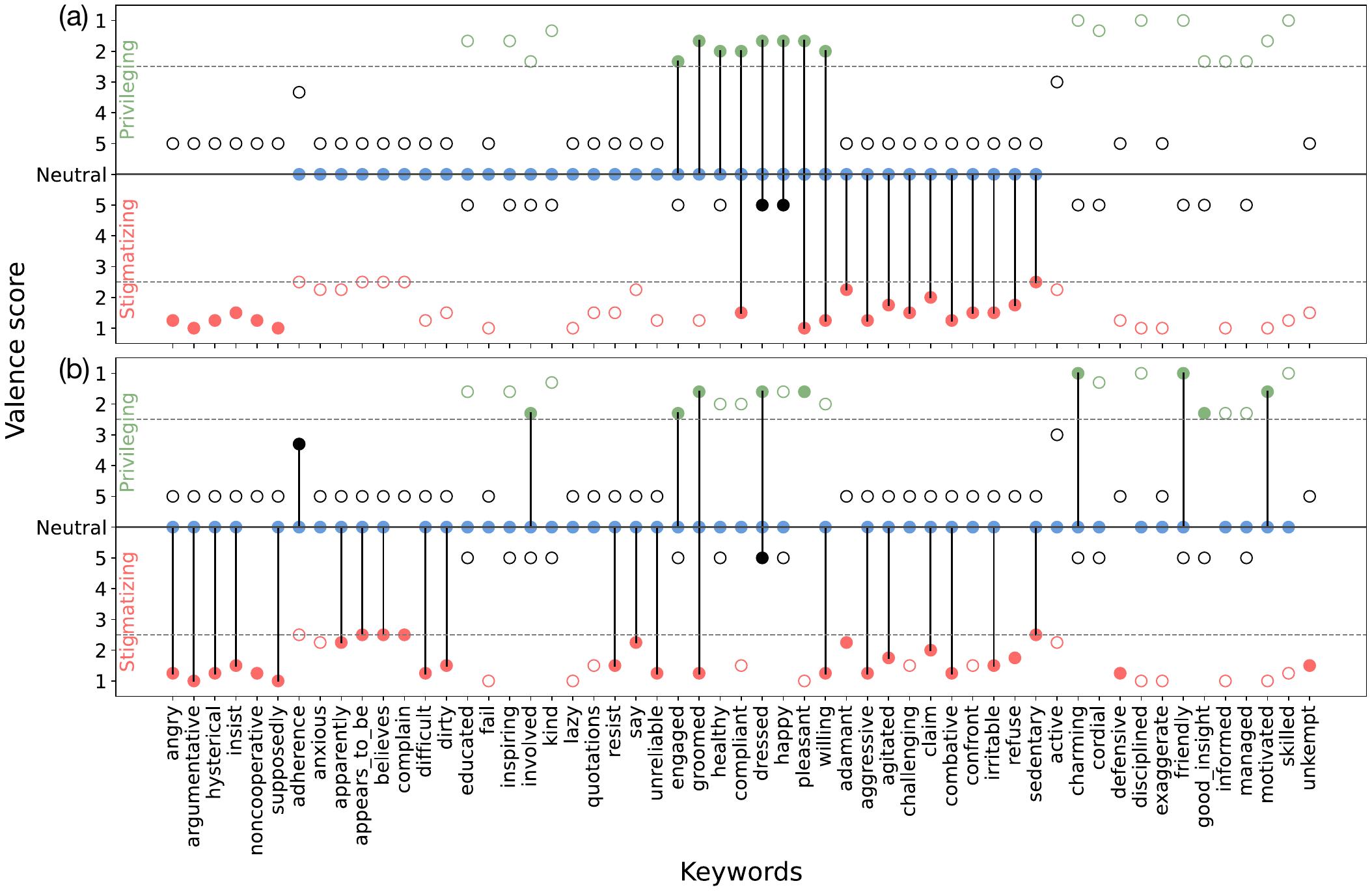} 
   \caption{Mean lexicon scores for stigmatizing and privileging valence are shown alongside emotional valence annotation labels for OB-GYN (panel a) and MIMIC-IV (panel b) text chunks. Solid dots represent annotated valence labels, while vertical lines indicate the range of emotional valences observed in clinical note chunks relative to the original lexicon scores. Blue dots denote instances labeled as neutral during real-world annotation. Solid black dots represent words that were not assigned a stigmatizing or privileging score in the lexicon but were labeled as such during annotation of real-world text chunks.}
   \label{fig:valence} 
 \end{figure}

Figure~\ref{fig:valence} shows the emotional valence scores and the annotation labels assigned to each of the $55$ keywords during the lexicon creation and chunk annotation steps. Each circle represents the mean score a keyword received based on annotator ratings. The horizontal dotted lines on each side of the solid middle line indicate the cutoff thresholds used to determine whether a word was attributed a stigmatizing or privileging valence. Dots falling below these thresholds are colored green (privileging valence) or red (stigmatizing valence), while dots above the thresholds are shown in black, indicating that the word was not assigned the corresponding valence in the lexicon. For instance, the term ``groomed'' is shown with both a stigmatizing (score $1.25$) and privileging (score $1.66$) valence.

The top panel (Figure~\ref{fig:valence}a) also displays the real-world emotional valence from the annotations of the text chunks extracted from OB-GYN clinical notes. Each solid point indicates that at least one chunk containing the corresponding keyword was annotated with that valence. For example, all instances including the word ``angry'' were annotated as stigmatizing. Blue solid dots represent instances where the keyword appeared in a context annotated as neutral, introducing a new valence not present in the original lexicon scoring. The presence of vertical lines connecting polar opposites indicates that multiple valences were seen in context, capturing the more nuanced emotional tone beyond strictly stigmatizing or privileging interpretations. As an example, the keyword ``happy'', which was assigned a privileging score of $1.6$ during the lexicon creation step, was included in instances also annotated as stigmatizing or neutral in the OB-GYN dataset. The instance ``She has no complaints and is happy to go home today'' was labeled as privileging, consistent with the original score. Conversely, ``Happy to cover regional analgesia'' was annotated as neutral, and ``Pt stated that she was just not happy of the room she was in last night'' was labeled as stigmatizing. Solid black dots indicate that, although the keyword was not assigned a specific valence during scoring, a polarized tone was identified during the annotation of at least one corresponding chunk. This variation illustrates how contextual usage can shift the perceived emotional valence beyond the initial lexicon-based assignment, highlighting the importance of a two-step process for vocabulary creation and annotation.

Figure~\ref{fig:valence}b mirrors the same structure but displays the emotional valence annotations for text chunks from the MIMIC-IV dataset, which includes emergency department and inpatient notes. While the underlying lexicon scores remain the same, the annotation results show a clear shift in the spectrum of valences assigned to some of the keywords. This shift likely reflects differences in clinical context and documentation practices between specialties and hospital systems. For instance, the term ``difficult'' was annotated as neutral in the OB-GYN dataset when used in a procedural context (``Pediatrics was in attendance due to a difficult extraction''), but it was labeled as stigmatizing in MIMIC-IV when referring to a patient’s behavior (``He was a difficult historian given underlying history of severe alcohol abuse''). Similarly, ``argumentative'' was considered neutral in MIMIC-IV when describing behavior reported secondhand (``His wife relates that he has been increasingly argumentative'') but was labeled as stigmatizing in OB-GYN when used to describe direct patient noncompliance (``At this time the patient is very argumentative and refusing c/s''). 

Overall, $54\%$ of OB-GYN instances and $60\%$ of MIMIC-IV instances were labeled as neutral. Neutral labels were typically assigned to text snippets that were contextually specific to the clinical specialty, conveyed information about individuals other than the patient, included language reported by a third party, or consisted of templates or administrative content (e.g., ``We are happy that you are feeling better, and it was a pleasure taking care of you!'' in MIMIC-IV). Among the remaining instances, OB-GYN text chunks were more often labeled as privileging ($26.9\%$) than stigmatizing ($18.7\%$), whereas the reverse trend was observed in MIMIC-IV, where $26.9\%$ of chunks were labeled stigmatizing and only $12.7\%$ privileging. These trends reinforce the influence of clinical specialty  on the expression and perception of emotional valence in clinical documentation.

\subsection{Identification of Emotional Valence in Clinical Text via Language Models}

Our results (see Table~\ref{tab:results}) indicate that prompting LLMs alone is insufficient for accurately classifying emotional valence in clinical text, even when the model has been exposed to clinical jargon. In the zero-shot setting, Llama's and Med42's performances increased from $F_1=0.34$ to $F_1=0.78$ and from $F_1=0.39$ to $F_1=0.77$, respectively, with minimal task-grounding, i.e., lexically primed inputs. When also adding optimized instructions and in-context examples via DSPy, the performance further increased to $F_1=0.84$ and $F_1=0.87$ on the test set. However, instruction fine-tuning of a smaller Llama model led to a $\sim10\%$ increase in performance on the test set ($F_1$ scores $0.95$ and $0.94$) with reduced data sensitivity (narrower confidence intervals) over the best prompting configuration.

When comparing LLMs to PLMs, GatorTron with lexically primed inputs and prompt-based fine-tuning achieved the highest overall performance, with an $F_1$ score of $0.96$ on the OB-GYN test set. While the performance difference between encoder-based and generative models fine-tuned on the task is marginal, GatorTron offers practical advantages: it is smaller and does not require prompt optimization to achieve strong results. In contrast, Llama’s performance gains depend on careful instruction tuning, prompt optimization, and access to sufficient computational resources, which may limit its portability and scalability.

Under the same input and fine-tuning conditions, GatorTron achieved slightly higher scores ($\sim1.1\%$ increase) compared to BERT, suggesting that exposure to real-world clinical narratives during pretraining may enhance the model’s ability to perform the task. However, this modest improvement—especially when contrasted with the larger gains observed through model fine-tuning and lexical priming—supports the conclusion that understanding how emotional valence is characterized in clinical text is more critical for identifying biased language than clinical domain knowledge alone.

We evaluated generalization performance across two axes: healthcare systems and medical specialties. Models fine-tuned on OB-GYN data showed limited generalization when validated on MIMIC-IV, with the best GatorTron configuration experiencing a $44\%$ drop in $F_1$ score (Table~\ref{tab:results}). This substantial performance degradation suggests that specialty-specific patterns make cross-domain transfer difficult. 

Reversing the training direction yielded markedly different results (see Table~\ref{tab:f1mimic}). The best model (prompt-tuned GatorTron, $F_1=0.80$) experienced only an $11\%$ performance drop when validated on OB-GYN data. Despite lower absolute performance, likely due to MIMIC-IV's smaller sample size for each specialty, training on heterogeneous medical specialties improved generalizability. 

The asymmetry in performance drops ($44\%$ vs. $11\%$) reveals that specialty diversity in training/evaluation data matters more than institutional differences. While models appear robust to distribution shifts across healthcare systems, they struggle with the semantic variability of emotional valence across medical specialties. A term considered neutral in one specialty may carry stigmatizing connotations in another, requiring specialty-specific adaptation for reliable bias detection.

In conclusion, our results demonstrate that, regardless of the language model used, lexically primed inputs and task-specific fine-tuning are essential to achieve optimal performance in emotional valence classification. This combination enables effective model adaptation not only to the task itself but also to the clinical context of interest, accounting for variations across medical specialties. Among the evaluated models, GatorTron emerged as the most promising candidate for implementing a scalable and portable system for bias identification in clinical text, given its strong performance, lightweight architecture, and minimal reliance on prompt optimization.




\begin{table}[h!]
\centering
\scriptsize
\caption{$F_1$ scores across different input types and tuning strategies. Numbers in bold represent the highest scores per group of models (encoder-only vs decoder-only) and datasets. Mean and $95\%$ confidence intervals obtained via bootstrapping on the test and external validation datasets are reported.}
\begin{tabular}{lllccc}
\toprule
\multirow{2}{*}{\textbf{Model}} & \multirow{2}{*}{\textbf{Input}} & \multirow{2}{*}{\textbf{Training Strategy}} & \multicolumn{2}{c}{\textbf{OB-GYN}} & \multirow{2}{*}{\textbf{MIMIC-IV}} \\
\cmidrule{4-5}
&&& \textbf{Dev} & \textbf{Test} & \\
\midrule
\multirow{3}{*}{GatorTron}&\multirow{2}{*}{Anchored}&Standard FT&$0.9531$& $0.9456\ (0.9069, 0.9755)$& $\mathbf{0.6765}\ (0.6133, 0.7323)$\\
&&Prompt FT&$0.9635$& $0.9487\ (0.9117, 0.9794)$& $0.5918\ (0.5281, 0.6560)$\\
\cmidrule{2-6}
&LP&Prompt FT&$0.9783$& $\mathbf{0.9572}\ (0.9250, 0.9839)$& $0.6598\ (0.5954, 0.7210)$\\
\midrule
 \multirow{3}{*}{BERT}&\multirow{2}{*}{Anchored}&Standard FT& $0.9510$& $0.9154\ (0.8695, 0.9540)$& $0.5910\ (0.5233, 0.6572)$\\
 &&Prompt FT& $0.9581$& $0.9311\ (0.8883, 0.9665)$& $0.5655\ (0.4967, 0.6341)$\\
 \cmidrule{2-6}
 &LP&Prompt FT& $0.9681$& $0.9461\ (0.9093, 0.9753)$& $0.5932\ (0.5243, 0.6639)$\\
\midrule
\midrule
\multirow{3}{*}{Llama-3.3-70B}&Anchored&\multirow{2}{*}{Zero-shot}&\multicolumn{2}{c}{$0.3387\ (0.2946, 0.3825)$}&$0.3512\ (0.3005, 0.4106)$\\
&LP&&\multicolumn{2}{c}{$0.7783\ (0.7361, 0.8215)$}&$0.5993\ (0.5406, 0.6588)$\\
\cmidrule{2-6}
&LP&Optimized ICL&$0.8440$& $0.8354\ (0.7772, 0.8886)$ & $0.5915\ (0.5283, 0.6513)$\\
\midrule
\multirow{2}{*}{Llama-3.2-1B}&ICL + LP&\multirow{2}{*}{Instruction FT}&$0.9725$& $0.9475\ (0.9127, 0.9774)$& $0.5083\ (0.4364, 0.5739)$\\
&LP&&$0.9497$& $\mathbf{0.9525}\ (0.9213, 0.9813)$& $0.5572\ (0.4896, 0.6237)$\\
\midrule
\multirow{3}{*}{Med42-70B}&Anchored&\multirow{2}{*}{Zero-shot}&\multicolumn{2}{c}{$0.3858\ (0.3339; 0.4352)$}&$0.4422\ (0.3720; 0.5146)$\\
&LP&&\multicolumn{2}{c}{$0.7242\ (0.6786; 0.7702)$}&$0.5871\ (0.5290; 0.6411)$\\
\cmidrule{2-6}
&LP&Optimized ICL&$0.8343$&$0.8695\ (0.8188; 0.9167)$&$\mathbf{0.6179}\ (0.5625; 0.6742)$\\
\midrule
\multirow{2}{*}{Med42-8B}&ICL + LP&\multirow{2}{*}{Instruction FT}&$0.9655$&$0.9399\ (0.9002; 0.9721)$&$0.5836\ (0.5162; 0.6565)$\\
&LP&&$0.9420$&$0.9155\ (0.8673; 0.9512)$&$0.5239\ (0.4590; 0.5895)$\\
\bottomrule
\end{tabular}
\label{tab:results}
\vspace{-0.9em} 
\begin{flushleft}
    \scriptsize LP: Lexically Primed; FT: Fine-Tuning; ICL: In-Context Learning.
\end{flushleft}
\end{table}

\section{Discussion} 
In this work, we showed that stigmatizing and privileging terms can elicit heterogeneous emotional responses in real-world clinical contexts, with the same term often appearing in stigmatizing, privileging, and neutral instances. Moreover, the extent to which terms are used in a polarized or neutral manner is closely tied to the medical specialty and institution in which the clinical note was written. Comparing annotations from OB-GYN notes from the Mount Sinai Healthcare system and inpatient and emergency department notes from MIMIC-IV, we found that the emotional valence of terms often shifted between the two datasets. In both settings, however, neutral valence was most frequently associated with administrative language, procedural descriptions, or references to individuals other than the patient.

To automatically capture this variability, we showed that both encoder-based models and generative LLMs require contextual adaptation through lexical priming and fine-tuning. While all fine-tuned models performed well on held-out data from the same medical specialties, performance dropped substantially when validated on different specialties. Additionally, using a language model pretrained on clinical text provided only moderate advantage over general-domain models. These results suggest that learning specialty-specific expressions of emotional valence is more critical than clinical pretraining or robustness to institutional differences alone.


Among the models evaluated, the encoder-based language model pretrained on clinical text (GatorTron) emerged as the best candidate for emotional valence classification in clinical settings via prompt-based fine-tuning and lexical priming. It achieved the highest overall performance, adapted effectively to clinical context with minimal prompt engineering, and offered greater scalability, with 345M parameters compared to the 1B parameters required by the smallest Llama model.

In conclusion, clinical note language is uniquely shaped by the medical setting and specialty, in that it serves multiple stakeholders, including billing specialists, legal representatives, other clinicians, and, following the 21st Century Cures Act\footnote{\url{https://www.fda.gov/regulatory-information/selected-amendments-fdc-act/21st-century-cures-act}}, the patients and their caregivers. This multi-purpose nature leads to the use of the same clinical terminologies in templates, standardized text for legal purposes, and narrative descriptions of the patient-clinician interactions. As a result, characterizing and extracting biased language becomes particularly challenging, as the same terms may convey heterogeneous and often difficult-to-label emotional valence.

Here, we advocate for the operationalization of biased language as the emotional valence elicited by clinical text, and its extraction using language models fine-tuned to the context of interest. When applied within intersectional frameworks, the identification and quantification of polarized language can help disentangle different social dynamics, on the one hand, strained therapeutic relationships and provider burnout, on the other, systemic bias and implicit assumptions about patients based on their sociodemographic characteristics. Making this distinction is critical to avoid conflating fundamentally different sources of biased language, and to more effectively guide efforts toward equitable care for minoritized and marginalized communities.

\paragraph{Limitations}
This study has limitations that warrant consideration. Our annotation framework captures emotional valence exclusively from the clinician’s perspective, which may overlook how language is perceived by patients or other stakeholders \citep{valentine2024}. Future work should incorporate diverse viewpoints to provide a more comprehensive understanding of how bias manifests in clinical text.

Moreover, our comparison across institutions is based on a single specialty (OB-GYN) in one system and a mixture of specialties in MIMIC-IV. Including additional specialties within the same healthcare system would allow us to better understand the sources of distribution shift in emotional valence.


\bibliography{sample}

\newpage
\appendix
\section*{Appendix A. Stigmatizing/Privileging Lexicon and Chunks}
\begin{table}[h!]
\centering
\small
\caption{Examples of lexicon and sentences for stigmatizing and privileging valence scoring.}
\begin{tabular}{lll}
\toprule
\textbf{Valence} & \textbf{Word(s)} & \textbf{Example}\\
\midrule
\multirow{3}{*}{Privileging} & engaged, engagement, engages & \emph{Patient is highly \underline{engaged} in their care.}\\
& pleasant & \emph{Patient was \underline{pleasant} throughout encounter.}\\
& compliant & \emph{Patient is \underline{compliant} with all recommendations.}\\
\midrule
\multirow{4}{*}{Stigmatizing} & noncompliant, noncompliance, & \emph{...has been \underline{noncompliant} with follow up.}\\
& poor compliance & \emph{\underline{Noncompliant} with treatment recommendations.}\\
& lazy & \emph{Patient seems too \underline{lazy} to make lifestyle changes.}\\
& irritable & \emph{Patient was \underline{irritable} during interview/exam.}\\
\bottomrule
\end{tabular}
\label{tab:a1}
\end{table}

\begin{table}[h!]
\centering
\small
\caption{Clinical note chunks labeled as neutral, privileging, or stigmatizing from the OB-GYN and MIMIC-IV datasets.}
\begin{tabular}{llp{10cm}}
\toprule
\textbf{Dataset} & \textbf{Annotation} & \textbf{Text Chunk}\\
\midrule
\multirow{6}{*}{OB-GYN} & \multirow{2}{*}{Neutral} & \emph{...Mrs. \_\_\_ is \underline{unwilling} to divulge the patient's psychiatric history...}\\
& \multirow{2}{*}{Privileging} & \emph{...PO intake gradually increasing; more \underline{willing} to take PO psychiatric medications...}\\
& \multirow{2}{*}{Stigmatizing} & \emph{...Pt is \underline{unwilling} to continue her IOL She feels that she has been in pain since hours prior and wants to be delivered...}\\
\midrule
\multirow{6}{*}{MIMIC-IV} & \multirow{2}{*}{Neutral} & \emph{...You are aware that there are food services such as Wheels on Meals, if your father is \underline{willing}...}\\
& \multirow{2}{*}{Privileging} & \emph{...Upon arrival to the floor, the patient is \underline{pleasant} and in no distress...}\\
& \multirow{2}{*}{Stigmatizing} & \emph{...He has been \underline{unwilling} to eat or get out of bed lately as per the daughter...}\\
\bottomrule
\end{tabular}
\label{tab:a2}
\end{table}

\section*{Appendix B. Fine-Tuning Details for GatorTron and BERT Models}
\begin{table}[h!]
\centering
\small
\caption{Single- and multi-word verbalizer mappings.}
\begin{tabular}{lll}
\toprule
\textbf{Method} & \textbf{Word(s)} & \textbf{Target Class}\\
\midrule
\multirow{3}{*}{Single-Word} & negative & stigmatizing\\
& positive & privileging\\
& neutral & neutral\\
\midrule
\multirow{3}{*}{Multi-Word} & negative, stigma, bad & stigmatizing\\ 
& positive, privilege, good & privileging\\
& neutral, instruction, clinical & neutral\\
\bottomrule
\end{tabular}
\label{tab:b1}
\end{table}

Hyperparameter tuning was done exploring all possible combinations of learning rate (range $[1\cdot10^{-5}; 2.5\cdot10^{-5}]$), verbalizer (single-word vs. multi-word), and batch size (set $\{1, 4, 8, 16\}$). Optimal hyperparameter combinations were: learning rate $1\cdot10^{-5}$ and batch size $8$ for standard fine-tuning; learning rate $2.5\cdot10^{-5}$, batch size $1$, and single-word verbalizer for prompt-based fine-tuning without lexical priming; and learning rate $2.5\cdot10^{-5}$, batch size of $8$, and single-word verbalizer for prompt-based fine-tuning with lexical priming.

\section*{Appendix C. Prompting and Fine-Tuning Details for Llama Models}
We optimized Llama 3.3 and Med42-70B versions using MIPROv2, on either OB-GYN or MIMIC-IV datasets, varying the number of evaluation trials and example candidates, mini-batch size, and development set size according to the three standard configurations available in DSPy, i.e., \emph{light}, \emph{medium}, and \emph{heavy}. The best configuration, as evaluated on the development set, was the medium for both optimizations of Llama 3.3, with $25$ trials, mini-batch size equal $25$, $17$ candidate examples, and the full development set for evaluation. The best configuration for Med42 optimized on the OB-GYN dataset was the \emph{medium} (same hyperparameters as above), whereas it was the \emph{heavy} when optimized on the MIMIC-IV dataset, with $50$ trials, mini-batch size equal $25$, $38$ candidate examples, and the full development set for evaluation. See Table~\ref{tab:c1} for the overview of the instructions considered.

\begin{table}[h!]
\scriptsize
\centering
\caption{Instructions for generative tasks with large language models. Optimized instructions are derived from tuning the prompts on (1) OB-GYN; and (2) MIMIC-IV datasets.}
\begin{tabular}{p{3cm}p{3cm}lp{6cm}}
\toprule
\textbf{Input} & \textbf{Tuning Strategy} & \textbf{Writer} & \textbf{Instruction} \\
\midrule
Anchored chunks & Zero-shot & Human & \emph{Classify whether the chunk extracted from a clinical note carries a privileging, stigmatizing, or neutral valence.} \\
LP chunks & Zero-shot & Human &\emph{Classify whether the word indicated at the beginning of the chunk extracted from a clinical note carries a privileging, stigmatizing, or neutral valence in the context provided.} \\
\midrule
LP chunks& MIPROv2& Llama 3.3-70B& 1. \emph{Analyze the given clinical note chunk, focusing on the word indicated at the beginning, and classify its valence as privileging, stigmatizing, or neutral based on the context provided. Consider the emotional tone conveyed by the language used in the chunk, taking into account the medical terminology, patient description, and overall sentiment expressed. Assign a valence that accurately reflects the emotional connotation of the word within the clinical note, ensuring that your classification is informed by the nuances of medical communication and the specific details presented in the chunk.}\\
&&& 2. \emph{Analyze the given chunk of text from a clinical note, focusing on the word indicated at the beginning. Determine whether this word conveys a privileging, stigmatizing, or neutral valence within the provided context, and classify it accordingly.} \\
\midrule
LP chunks & MIPROv2& Med42-70B&1. \emph{For each identified keyword in a clinical note chunk (e.g., "pleasant," "adherence"), determine its valence as privileging, stigmatizing, or neutral by considering both medical context and linguistic cues. Implement domain-specific sentiment analysis rules or train a machine learning model on annotated datasets to assign accurate valences. Ensure the instruction emphasizes capturing nuanced expressions typical in clinical documentation while prioritizing factual representations over emotional bias.}\\
&&& \emph{2. In a critical care setting, accurately classify sentiment valence (privileging, stigmatizing, or neutral) in clinical text chunks to inform timely interventions and patient summaries. Given phrases from psychiatric and internal medicine notes, such as "difficulty inspiring" for respiratory symptoms or "appropriately dressed" for mental status evaluation, the Language Model must assign correct valence labels considering domain-specific context.}\\
\bottomrule
\end{tabular}
\label{tab:c1}  
\vspace{-0.9em} 
\begin{flushleft}
    \scriptsize LP: Lexically Primed
\end{flushleft}
\end{table}

The instruction fine-tuning of the Llama 3.2 1B and Med42-8B models was performed on both OB-GYN and MIMIC-IV training sets over $10$ epochs with early stopping criteria with patience equal $3$. We instruction fine-tuned Med42-8B with LoRA and gradient accumulation steps equal $4$. Training was conducted with fully sharded data parallelism on two H100 GPUs.
In all configurations, learning rates varied from $1\cdot10^{-6}$ to $1\cdot10^{-4}$ for OB-GYN and from $1\cdot10^{-8}$ to $1\cdot10^{-4}$ for MIMIC-IV. 

When fine-tuning on OB-GYN, best learning rates for the the models with optimized instructions and few-shot examples, based on the $F_1$ score obtained on the development set, were $1\cdot10^{-6}$ and $1\cdot10^{-4}$, respectively. Whereas for the models with only the optimized instruction the best learning rates were $1\cdot10^{-5}$ and $5\cdot10^{-4}$, respectively. Conversely, for MIMIC-IV, best learning rate for both input configurations was $1\cdot10^{-6}$ for Llama 3.3 1B and $1\cdot10^{-4}$ and $5\cdot10^{-5}$, respectively, for Med42 8B.

We set random seed equal $42$ and temperature equal $0$ for replicable output generations in all experiments.

\section*{Appendix D. Models Performance}

\begin{table}[h!]
\centering
\scriptsize
\caption{Precision scores across different input types and tuning strategies. Mean and $95\%$ confidence intervals obtained via bootstrapping on the test and external validation datasets are reported.}
\begin{tabular}{lllccc}
\toprule
\multirow{2}{*}{\textbf{Model}} & \multirow{2}{*}{\textbf{Input}} & \multirow{2}{*}{\textbf{Training Strategy}} & \multicolumn{2}{c}{\textbf{OB-GYN}} & \multirow{2}{*}{\textbf{MIMIC-IV}} \\
\cmidrule{4-5}
&&& \textbf{Dev} & \textbf{Test} & \\
\midrule
\multirow{3}{*}{GatorTron}&\multirow{2}{*}{Anchored}&Standard FT & $0.9481$ & $0.9361\ (0.8892, 0.9720)$ & $0.6704\ (0.5986, 0.7348)$ \\
&&Prompt FT & $0.9577$ & $0.9388\ (0.8944, 0.9762)$ & $0.5998\ (0.5354, 0.6671)$\\
\cmidrule{2-6}
&LP&Prompt FT& $0.9745$ & $0.9503\ (0.9111, 0.9818)$ & $0.6480\ (0.5824, 0.7155)$\\
\midrule
 \multirow{3}{*}{BERT}&\multirow{2}{*}{Anchored}&Standard FT& $0.9520$&  $0.9193\ (0.8714, 0.9599)$ & $0.5982\ (0.5256, 0.6700)$ \\
 &&Prompt FT& $0.9481$& $0.9172\ (0.8683, 0.9603)$ & $0.5618\ (0.4935, 0.6356)$ \\
 \cmidrule{2-6}
 &LP&Prompt FT& $0.9637$& $0.9432\ (0.9044, 0.9760)$& $0.6068\ (0.5327, 0.6804)$\\
\midrule
\midrule
\multirow{3}{*}{Llama-3.3-70B}&Anchored&\multirow{2}{*}{Zero-shot}&\multicolumn{2}{c}{$0.4220\ (0.3539, 0.4865)$} & $0.4938\ (0.3020, 0.7156)$\\
&LP&&\multicolumn{2}{c}{$0.7611\ (0.7172, 0.8059)$} & $0.5739\ (0.5155, 0.6347)$\\
\cmidrule{2-6}
&LP&Optimized ICL& $0.8378$ & $0.8376\ (0.7767, 0.8916)$ & $0.5725\ (0.5107, 0.6319)$\\
\midrule
\multirow{2}{*}{Llama-3.2-1B}&ICL + LP&\multirow{2}{*}{Instruction FT}& $0.9710$& $0.9451\ (0.9049, 0.9770)$& $0.5497\ (0.4605, 0.6334)$\\
&LP& & $0.9427$& $0.9444\ (0.9035, 0.9814)$& $0.6020\ (0.5182, 0.6836)$\\
\midrule
\multirow{3}{*}{Med42-70B}&Anchored&\multirow{2}{*}{Zero-shot}&\multicolumn{2}{c}{$0.6189\ (0.4910; 0.7498)$}&$0.5486\ (0.4286; 0.6704)$\\
&LP&&\multicolumn{2}{c}{$0.7147\ (0.6720; 0.7576)$}&$0.5672\ (0.5103; 0.6223)$\\
\cmidrule{2-6}
&LP&Optimized ICL&$0.8188$&$0.8510\ (0.7946; 0.9028)$&$0.5961\ (0.5442; 0.6528)$\\
\midrule
\multirow{2}{*}{Med42-8B}&ICL + LP&\multirow{2}{*}{Instruction FT}&$0.9605$&$0.9469\ (0.9051; 0.9770)$&$0.64168\ (0.5603;  0.7242)$\\
&LP&&$0.9363$&$0.9099\ (0.8586; 0.9514)$&$0.5380\ (0.4675; 0.6079)$\\

\bottomrule
\end{tabular}
\label{tab:resultsPrecision}
\vspace{-0.9em} 
\begin{flushleft}
    \scriptsize LP: Lexically Primed; FT: Fine-Tuning; ICL: In-Context Learning.
\end{flushleft}
\end{table}

\begin{table}[h!]
\centering
\scriptsize
\caption{Recall scores across different input types and tuning strategies. Mean and $95\%$ confidence intervals obtained via bootstrapping on the test and external validation datasets are reported.}
\begin{tabular}{lllccc}
\toprule
\multirow{2}{*}{\textbf{Model}} & \multirow{2}{*}{\textbf{Input}} & \multirow{2}{*}{\textbf{Training Strategy}} & \multicolumn{2}{c}{\textbf{OB-GYN}} & \multirow{2}{*}{\textbf{MIMIC-IV}} \\
\cmidrule{4-5}
&&& \textbf{Dev} & \textbf{Test} & \\
\midrule
\multirow{3}{*}{GatorTron}&\multirow{2}{*}{Anchored}&Standard FT & $0.9585$ & $0.9576\ ( 0.9245, 0.9817)$ & $0.7115\ (0.6547, 0.7654)$\\
&&Prompt FT & $0.9699$ & $0.9637\ (0.9365, 0.9855)$ & $0.6042\ (0.5371, 0.6727)$\\
\cmidrule{2-6}
&LP&Prompt FT& $0.9823$ & $0.9662\ (0.9384, 0.9885)$ & $0.6808\ (0.6138, 0.7431)$ \\
\midrule
 \multirow{3}{*}{BERT}&\multirow{2}{*}{Anchored}&Standard FT& $0.9502$ & $0.9132\ (0.8624, 0.9526)$ & $0.6016\ (0.5319, 0.6673)$ \\
 &&Prompt FT& $0.9699$ & $0.9504\ (0.9156, 0.9777)$& $0.5852\ (0.5164, 0.6529)$ \\
 \cmidrule{2-6}
 &LP&Prompt FT& $0.9730$ & $0.9506\ (0.9129, 0.9795)$ & $0.5945\ (0.5217, 0.6601)$ \\
\midrule
\midrule
\multirow{3}{*}{Llama-3.3-70B}&Anchored&\multirow{2}{*}{Zero-shot}&\multicolumn{2}{c}{$0.3894\ (0.3591, 0.4204)$} & $0.3700\ (0.3330, 0.4119)$\\
&LP&&\multicolumn{2}{c}{$0.8146\ (0.7765, 0.8537)$} & $0.6660\ (0.6087, 0.7234)$\\
\cmidrule{2-6}
&LP&Optimized ICL& $0.8547$ & $0.8532\ (0.7960, 0.9035)$ & $0.6825\ (0.6366, 0.7298)$\\
\midrule
\multirow{2}{*}{Llama-3.2-1B}&ICL + LP&\multirow{2}{*}{Instruction FT}& $0.9740$& $0.9510\ (0.9183, 0.9792)$& $0.4936\ (0.4299, 0.5572)$\\
&LP& & $0.9575$& $0.9635\ (0.9379, 0.9854)$& $0.5388\ (0.4742, 0.6038)$\\
\midrule
\multirow{3}{*}{Med42-70B}&Anchored&\multirow{2}{*}{Zero-shot}&\multicolumn{2}{c}{$0.4154\ (0.3802; 0.4540)$}&$0.4280\ (0.3711; 0.4860)$\\
&LP&&\multicolumn{2}{c}{$0.7818\ (0.7409; 0.8212)$}&$0.6784\ (0.6247; 0.7294)$\\
\cmidrule{2-6}
&LP&Optimized ICL&$0.8733$&$0.9050\ (0.8598; 0.9437)$&$0.7190\ (0.6729; 0.7637)$\\
\midrule
\multirow{2}{*}{Med42-8B}&ICL + LP&\multirow{2}{*}{Instruction FT}&$0.9709$&$0.9352\ (0.8903; 0.9710)$&$0.5623\ (0.4962; 0.6316)$\\
&LP&&$0.9483$&$0.9232\ (0.8797; 0.9578)$&$0.5369\ (0.4685; 0.6040)$\\
\bottomrule
\end{tabular}
\label{tab:resultsRecall}
\vspace{-0.9em} 
\begin{flushleft}
    \scriptsize LP: Lexically Primed; FT: Fine-Tuning; ICL: In-Context Learning.
\end{flushleft}
\end{table}

\begin{table}[h!]
\centering
\scriptsize
\caption{$F_1$ scores across different input types and tuning strategies when training on MIMIC-IV and testing on OB-GYN datasets. Mean and 95\% confidence intervals obtained via bootstrapping on the test and external validation datasets are reported.}
\begin{tabular}{lllccc}
\toprule
\multirow{2}{*}{\textbf{Model}} & \multirow{2}{*}{\textbf{Input}} & \multirow{2}{*}{\textbf{Training Strategy}} & \multicolumn{2}{c}{\textbf{MIMIC-IV}} & \multirow{2}{*}{\textbf{OB-GYN}} \\
\cmidrule{4-5}
&&& \textbf{Dev} & \textbf{Test} & \\
\midrule
\multirow{3}{*}{GatorTron}&\multirow{2}{*}{Anchored}&Standard FT & $0.6168$ & $0.5949\ (0.4187, 0.7640)$  & $0.5901\ (0.5554, 0.6247)$ \\
&&Prompt FT & $0.7532$ & $0.8034\ (0.6482, 0.9136)$ & $0.7144\ (0.6843, 0.7455)$ \\
\cmidrule{2-6}
&LP&Prompt FT& $0.9381$ & $0.7189\ (0.5477, 0.8619)$  & $0.6547\ (0.6225, 0.6853)$ \\
\midrule
 \multirow{3}{*}{BERT}&\multirow{2}{*}{Anchored}&Standard FT& $0.5166$ & $0.5190\ (0.3552, 0.6863)$  & $0.4773\ (0.4452, 0.5102)$ \\
 &&Prompt FT& $0.7315$ & $0.7786\ (0.6249, 0.9010)$ & $0.6201\ (0.5859, 0.6521)$ \\
 \cmidrule{2-6}
 &LP&Prompt FT& $0.9122$ &  $0.6582\ (0.4979, 0.8025)$ & $0.5909\ (0.5577, 0.6218)$ \\
\midrule
\midrule
Llama-3.3-70B&LP&Optimized ICL& $0.6936$& $0.5488\ (0.3955, 0.6983)$ &$0.6368\ (0.6028; 0.6704)$ \\
\midrule
\multirow{2}{*}{Llama-3.2-1B}&ICL + LP&\multirow{2}{*}{Instruction FT}& $0.8536$&  $0.6803\ (0.5322; 0.8199)$& $0.5422\ (0.5079; 0.5758)$\\
&LP& & $0.8725$&  $0.7856\ (0.6302; 0.9128)$& $0.7334\ (0.7009; 0.7628)$\\
\midrule
Med42-70B&LP&Optimized ICL& $0.7100$& $0.5713\ (0.4169, 0.7094)$ & $0.7709\ (0.7449; 0.7960)$\\
\midrule
\multirow{2}{*}{Med42-8B}&ICL + LP&\multirow{2}{*}{Instruction FT}&$0.7762$&$0.6460\ (0.6104; 0.6781)$&$0.6546\ (0.4940; 0.7999)$\\
&LP& &$0.6958$ &$0.5595\ (0.4122; 0.6989)$&$0.8197\  (0.7948; 0.8435)$\\
\bottomrule
\end{tabular}
\label{tab:f1mimic}
\vspace{-0.9em} 
\begin{flushleft}
    \scriptsize LP: Lexically Primed; FT: Fine-Tuning; ICL: In-Context Learning.
\end{flushleft}
\end{table}

\begin{table}[h!]
\centering
\scriptsize
\caption{Precision scores across different input types and tuning strategies when training on MIMIC-IV and testing on OB-GYN datasets. Mean and 95\% confidence intervals obtained via bootstrapping on the test and external validation datasets are reported.}
\begin{tabular}{lllccc}
\toprule
\multirow{2}{*}{\textbf{Model}} & \multirow{2}{*}{\textbf{Input}} & \multirow{2}{*}{\textbf{Training Strategy}} & \multicolumn{2}{c}{\textbf{MIMIC-IV}} & \multirow{2}{*}{\textbf{OB-GYN}} \\
\cmidrule{4-5}
&&& \textbf{Dev} & \textbf{Test} & \\
\midrule
\multirow{3}{*}{GatorTron}&\multirow{2}{*}{Anchored}&Standard FT & $0.6741$ & $0.6147\ (0.4324, 0.8046)$  & $0.6839\ (0.6452, 0.7230)$ \\
&&Prompt FT & $0.7468$ & $0.7755\ (0.6258, 0.9020)$   & $0.7748\ (0.7461, 0.8022)$ \\
\cmidrule{2-6}
&LP&Prompt FT& $0.9484$ & $0.7185\ (0.5555, 0.8693)$  & $0.7680\ (0.7399, 0.7957)$ \\
\midrule
 \multirow{3}{*}{BERT}&\multirow{2}{*}{Anchored}&Standard FT&$0.5748$& $0.5306\ (0.3659, 0.7089)$ & $0.5660\ (0.5299, 0.6029)$ \\
 &&Prompt FT& $0.7349$ & $0.7998\ (0.6227, 0.9350)$  & $0.7224\ (0.6924, 0.7516)$ \\
 \cmidrule{2-6}
 &LP&Prompt FT& $0.9076$ & $0.6461\ (0.5000, 0.7954)$ & $0.6676\ (0.6320, 0.7010)$ \\
\midrule
\midrule
Llama-3.3-70B&LP&Optimized ICL& $0.7121$&  $0.5555\ (0.3965, 0.7125)$&$0.7146\ (0.6769; 0.7486)$ \\
\midrule
\multirow{2}{*}{Llama-3.2-1B}&ICL + LP&\multirow{2}{*}{Instruction FT}& $0.8503$&  $0.6608\ (0.5244; 0.8056)$&$0.5959\ (0.5617; 0.6303)$ \\
&LP& & $0.9043$&  $0.8185\ (0.6646; 0.9445)$& $0.8195\ (0.7907; 0.8466)$\\
\midrule
Med42-70B&LP&Optimized ICL& $0.6840$&$0.5601\ (0.4270, 0.6906)$& $0.7557\ (0.7302; 0.7824)$\\
\midrule
\multirow{2}{*}{Med42-8B}&ICL + LP&\multirow{2}{*}{Instruction FT}& $0.8372$&$0.7681\ (0.7378; 0.7937)$&$0.6462\ (0.4839; 0.7962)$ \\
&LP& & $0.6832$&$0.5427\ (0.4108; 0.6729)$& $0.8131\ (0.7872; 0.8385)$\\
\bottomrule
\end{tabular}
\label{tab:resultsP}
\vspace{-0.9em} 
\begin{flushleft}
    \scriptsize LP: Lexically Primed; FT: Fine-Tuning; ICL: In-Context Learning.
\end{flushleft}
\end{table}

\begin{table}[h!]
\centering
\scriptsize
\caption{Recall scores across different input types and tuning strategies when training on MIMIC-IV and testing on OB-GYN datasets. Mean and 95\% confidence intervals obtained via bootstrapping on the test and external validation datasets are reported.}
\begin{tabular}{lllccc}
\toprule
\multirow{2}{*}{\textbf{Model}} & \multirow{2}{*}{\textbf{Input}} & \multirow{2}{*}{\textbf{Training Strategy}} & \multicolumn{2}{c}{\textbf{MIMIC-IV}} & \multirow{2}{*}{\textbf{OB-GYN}} \\
\cmidrule{4-5}
&&& \textbf{Dev} & \textbf{Test} & \\
\midrule
\multirow{3}{*}{GatorTron}&\multirow{2}{*}{Anchored}&Standard FT & $0.6278$ & $0.6027\ (0.4271, 0.7927)$  & $0.5662\ (0.5353, 0.5961)$ \\
&&Prompt FT & $0.7722$ & $0.8699\ (0.7756, 0.9482)$ & $0.6931\ (0.6617, 0.7257)$ \\
\cmidrule{2-6}
&LP&Prompt FT& $0.9308$ & $0.7428\ (0.5679, 0.8926)$ & $0.6328\ (0.6017, 0.6627)$ \\
\midrule
 \multirow{3}{*}{BERT}&\multirow{2}{*}{Anchored}&Standard FT& $0.4884$ & $0.5309\ (0.3629, 0.7100)$ & $0.4790\ (0.4502, 0.5076)$\\
 &&Prompt FT& $0.7350$ &  $0.7784\ (0.6049, 0.9078)$ &  $0.6075\ (0.5759, 0.6384)$ \\
 \cmidrule{2-6}
 &LP&Prompt FT& $0.9191$ & $0.7065\ (0.5249, 0.8569)$ & $0.5825\ (0.5502, 0.6137)$ \\
\midrule
\midrule
Llama-3.3-70B&LP&Optimized ICL& $0.7414$&  $0.6224\ (0.4623, 0.7602)$&$0.6125\ (0.5812; 0.6434)$ \\
\midrule
\multirow{2}{*}{Llama-3.2-1B}&ICL + LP&\multirow{2}{*}{Instruction FT}& $0.8591$ &  $0.7687\ (0.6114; 0.8955)$& $0.5044\ (0.4727; 0.5353)$\\
&LP& & $0.8719$&  $0.7914\ (0.6412; 0.9205)$& $0.5346\ (0.5027; 0.5672)$\\
\midrule
Med42-70B&LP&Optimized ICL& $0.7874$ & $0.7110\ (0.5935, 0.8176)$& $0.8293\ (0.8067; 0.8506)$\\
\midrule
\multirow{2}{*}{Med42-8B}&ICL + LP&\multirow{2}{*}{Instruction FT}&$0.7467$ &$ 0.6140\ ( 0.5806; 0.6445)$&$0.7035\ (0.5460; 0.8350)$\\
&LP& & $0.7130$&$0.6771\ (0.5698; 0.7833)$&$0.8338\ (0.8083; 0.8583)$\\
\bottomrule
\end{tabular}
\label{tab:resultsRecall}
\vspace{-0.9em} 
\begin{flushleft}
    \scriptsize LP: Lexically Primed; FT: Fine-Tuning; ICL: In-Context Learning.
\end{flushleft}
\end{table}

\section*{Appendix E. Error Analysis}
 We started by considering the keywords that matched only in the MIMIC-IV dataset ($N = 13$). For $7$ of these keywords, the GatorTron model correctly classified the extracted text. For the remaining $6$, misclassifications frequently involved content from psychiatric notes, suggesting that errors were driven by specialty-specific language patterns not encountered in the OB-GYN training data.

We then analyzed cases where keywords appeared in both datasets but carried different valences. For example, ``angry'' was labeled only as stigmatizing in OB-GYN but as both neutral and stigmatizing in MIMIC-IV. Similarly, ``believes'' was consistently labeled neutral in OB-GYN but also appeared as stigmatizing in MIMIC-IV. In these cases, the model often failed to correctly assign the new valence observed in MIMIC-IV. When errors did not involve unseen valence shifts, they typically occurred in chunks from psychiatric evaluations, further suggesting that specialty-specific language contributed to model failures.

Although GatorTron was pretrained on MIMIC-IV notes, which could partially explain its performance on chunks not seen during training, the misclassification of text related to psychiatric evaluations suggests that the observed drop in generalization may be more attributable to differences in clinical specialty than to institutional setting. Further investigation is warranted to better understand how specialty-specific language influences model performance across domains.

\end{document}